\documentclass{article}

\usepackage{amsmath}
\usepackage{color}
\usepackage{soul}
\usepackage{xspace}
\usepackage{dsfont}
\usepackage{bm}
\usepackage{bbm}
\usepackage{nicefrac}  %
\usepackage{amsfonts}
\usepackage{amssymb}
\usepackage{array}
\usepackage{mathtools}
\usepackage{caption}
\usepackage{subcaption}
\usepackage{fontawesome}
\usepackage{thmtools}
\usepackage{thm-restate}
\usepackage{lipsum}
\usepackage{pifont}
\usepackage{makecell}
\usepackage{multirow}
\usepackage{afterpage}
\usepackage{tabularx} 
\usepackage{url}
\usepackage[normalem]{ulem}
\usepackage{wrapfig}  %
\usepackage{textcomp}
\usepackage{amsmath,amssymb,amsfonts}
\usepackage{algorithmic}
\usepackage{graphicx}
\usepackage{textcomp}
\usepackage{algorithm}
\usepackage{wrapfig}
\usepackage{booktabs}
\usepackage{lipsum}
\usepackage{balance}
\usepackage{url}
\usepackage{amssymb}
\usepackage{amsmath}
\usepackage{amsfonts}       %
\usepackage{arydshln}
\usepackage{booktabs}       %
\usepackage{balance}
\usepackage{booktabs}
\usepackage[utf8]{inputenc} %
\usepackage[T1]{fontenc}    %
\usepackage{color, colortbl}
\usepackage{collcell}
\usepackage[pdftex, pdfstartview={FitV}, pdfpagelayout={TwoColumnLeft},bookmarksopen=true,plainpages = false, colorlinks=true, linkcolor=black, citecolor = black, urlcolor = black,filecolor=black , pagebackref=false,hypertexnames=false, plainpages=false, pdfpagelabels,bookmarks=false ]{hyperref}
\usepackage{url}            %
\usepackage{nicefrac}       %
\usepackage{microtype}      %
\usepackage{makecell}
\usepackage[font=footnotesize]{caption} %
\usepackage{subcaption}
\usepackage[]{siunitx}
\usepackage{graphicx}
\usepackage{epstopdf} %
\usepackage{mathtools}
\usepackage{marginnote}

\usepackage{multirow}
\usepackage[sort,compress]{cite}
\usepackage[dvipsnames, table]{xcolor} %

\usepackage{paralist}
\floatstyle{boxed} \floatname{algorithm}{Algorithm}
\newfloat{algorithm}{t}{loa}[section]
\usepackage{mathtools}

\usepackage{xstring}
\usepackage{color}
\usepackage{soul}
\usepackage{xspace}
\usepackage{dsfont}
\usepackage{bm}
\usepackage{bbm}
\usepackage{nicefrac}  %
\usepackage{amsfonts}
\usepackage{amssymb}
\usepackage{array}
\usepackage{mathtools}
\usepackage{subcaption}
\usepackage{fontawesome}
\usepackage{thmtools}
\usepackage{thm-restate}
\usepackage{lipsum}
\usepackage{pifont}
\usepackage{makecell}
\usepackage{multirow}
\usepackage{afterpage}
\usepackage{tabularx} 
\usepackage{textcase}
\usepackage{tikz}
\usepackage{url}
\usepackage[normalem]{ulem}
\usepackage{wrapfig}  %
\usepackage{textcomp}
\usepackage{xifthen}%

\let\labelindent\relax
\usepackage{enumitem}
\usepackage{verbatimbox}
\usepackage[breakable]{tcolorbox}
\usepackage{fancyvrb}
\usepackage{fancyref}
\usepackage{hyperref}
\usepackage{subcaption}
\usepackage[subtle,title=tight]{savetrees}
\usepackage{soul}

\usepackage{inconsolata}
\usepackage[T1]{fontenc}

\usepackage{pgffor}

\usepackage[]{algorithmic}
\usepackage{pgf}
\usepackage{makecell}
\usepackage{mathrsfs}
\usepackage[capitalize]{cleveref}

\definecolor{light-gray}{rgb}{0.8, 0.8, 0.8}
\definecolor{comment-green}{rgb}{0.435, 0.576, 0.106}
\definecolor{prompt-blue}{HTML}{2596be}
\definecolor{code-function}{HTML}{379fbe}
\definecolor{code-function}{HTML}{693da8}  %
\definecolor{code-syntax}{HTML}{0060b1}
\definecolor{code-constant}{HTML}{d86001}
\definecolor{prompt-gray}{HTML}{a7a7a7}
\definecolor{highlight}{HTML}{f8f9cb}
\definecolor{highlight}{HTML}{e3eeff}  %
\definecolor{code-perception}{HTML}{2ecc71}
\definecolor{code-control}{HTML}{ff9900}
\definecolor{code-undefined}{HTML}{ff0000}
\renewcommand\fbox{\fcolorbox{light-gray}{white}}

\newcommand{\vbf}[2][]{
    \ifthenelse{\isempty{#1}}%
    {\prescript{}{}{\mathbf{#2}}}%
    {\prescript{}{#1}{\mathbf{#2}}}%
}
\newcommand{\vbfd}[2][]{
    \ifthenelse{\isempty{#1}}%
    {\prescript{}{}{\dot{\mathbf{#2}}}}%
    {\prescript{}{\text{#1}}{\dot{\mathbf{#2}}}}%
}

\newif\ifcomments
\commentstrue

\ifcomments
	\newcommand{\aXX}[1]{\color{blue} AT: (#1)\color{black}\xspace}  %
	\newcommand{\XX}[1]{\color{red}JH: (#1)\color{black}\xspace}  %
\else
    \newcommand{\dXX}[1]{}  %
	\newcommand{\aXX}[1]{}  %
	\newcommand{\XX}[1]{}  %
\fi

\newcommand{\vbs}[2][]{
    \ifthenelse{\isempty{#1}}%
    {\prescript{}{}{\boldsymbol{#2}}}%
    {\prescript{}{\text{#1}}{\boldsymbol{#2}}}%
}
\newcommand{\vbsd}[2][]{
    \ifthenelse{\isempty{#1}}%
    {\prescript{}{}{\dot{\boldsymbol{#2}}}}%
    {\prescript{}{\text{#1}}{\dot{\boldsymbol{#2}}}}%
}

\newcommand{\usafe}{\bar{\mathbf{u}}}   %

\newcommand{\PreserveBackslash}[1]{\let\temp=\\#1\let\\=\temp}
\newcolumntype{C}[1]{>{\PreserveBackslash\centering}p{#1}}
\newcolumntype{R}[1]{>{\PreserveBackslash\raggedleft}p{#1}}
\newcolumntype{L}[1]{>{\PreserveBackslash\raggedright}p{#1}}

\definecolor{DarkGreen}{rgb}{0.0, 0.5, 0.0}

\definecolor{LightCyan}{rgb}{0.88,1,1}

\newcommand{\hlcoderesp}[1]{\colorbox{highlight}{\makebox(0.96\linewidth, 0.8\baselineskip)[l]{#1}}}

\NewDocumentCommand{\code}{v}{%
\texttt{\small{\textcolor{code-syntax}{#1}}}%
}

\newcommand{\query}[1]{\textcolor{comment-green}{#1}}
\newcommand{\prompt}[1]{\textcolor{prompt-gray}{#1}}
\newcommand{\resp}[1]{\textcolor{prompt-blue}{#1}}

\begingroup\lccode`~=`\_ \lowercase{\endgroup\let~}\_

\newcommand{\lmpchat}[2]{
\begin{tcolorbox}[boxsep=0pt,
                  left=3pt,
                  right=-4pt,
                  top=3pt,
                  bottom=3pt,
                  arc=0pt,
                  boxrule=0.5pt,
                  colframe=light-gray,
                  colback=white,
                  adjusted title=#1,
                  fonttitle=\small\ttfamily,
                  ]
\scriptsize{  %
\ttfamily
#2
}
\end{tcolorbox}
}

\newcommand{\Repeat}[2]{
    \foreach \n in {1,...,#1}{#2}
}

\newcommand\blfootnote[1]{%
  \begingroup
  \renewcommand\thefootnote{}\footnote{#1}%
  \addtocounter{footnote}{-1}%
  \endgroup
}
\usepackage[final]{corl_2023} %

\title{REAL: Resilience and Adaptation using Large Language Models  on Autonomous Aerial Robots}

\author{
Andrea Tagliabue$^{*}$\\
LIDS\\
MIT\\
\texttt{atagliab@mit.edu} \\
\And
Kota Kondo$^{*}$\\
LIDS\\
MIT\\
\texttt{kkondo@mit.edu}\\
\And
Tong Zhao$^{*}$\\
EECS\\
MIT\\
\texttt{tzhao@mit.edu}\\
\And
Mason Peterson$^{*}$\\
LIDS\\
MIT\\
\texttt{masonbp@mit.edu}\\
\And
Claudius T. Tewari \\
LIDS\\
MIT\\
\texttt{cttewari@mit.edu}\\
\And
Jonathan P. How\\
LIDS\\
MIT\\
\texttt{jhow@mit.edu}
}

\begin{document}
\maketitle
\blfootnote{*Equal Contribution. The authors are with the Department of Aeronautics and Astronautics, Massachusetts Institute of Technology. Work funded by the Air Force Office of Scientific Research MURI FA9550-19-1-0386. }%

\begin{abstract}
Large Language Models (LLMs) pre-trained on internet-scale datasets have shown impressive capabilities in code understanding, synthesis, and general purpose question-and-answering. 
Key to their performance is the substantial prior knowledge acquired during training and their ability to reason over extended sequences of symbols, often presented in natural language. 
In this work, we aim to harness the extensive long-term reasoning, natural language comprehension, and the available prior knowledge of LLMs for increased resilience and adaptation in autonomous mobile robots. 
We introduce \textbf{REAL}, an approach for \textbf{RE}silience and \textbf{A}daptation using \textbf{L}LMs. REAL provides a strategy to employ LLMs as a part of the mission planning and control framework of an autonomous robot. 
The LLM employed by REAL provides   
\begin{inparaenum}[(i)]
\item a source of prior knowledge to increase resilience for challenging scenarios that the system had not been explicitly designed for;
\item a way to interpret natural-language and other log/diagnostic information available in the autonomy stack, for mission planning;  
\item a way to adapt the control inputs using minimal user-provided prior knowledge about the dynamics/kinematics of the robot.
\end{inparaenum}
We integrate REAL in the autonomy stack of a real multirotor, querying onboard an offboard LLM at $0.1$-$1.0$ Hz as part the robot's mission planning and control feedback loops. We demonstrate in real-world experiments the ability of the LLM to reduce the position tracking errors of a multirotor under the presence of 
\begin{inparaenum}[(i)]
\item errors in the parameters of the controller and 
\item unmodeled dynamics. We also show
\item decision making to avoid potentially dangerous scenarios (e.g., robot oscillates) that had not been explicitly accounted for in the initial prompt design.
\end{inparaenum}
\end{abstract}


\section{Introduction}
Creating mission planning and control capabilities that are adaptive and resilient to unexpected scenarios has been a large area of research in recent years. Adaptive control has enabled exceptional performance when addressing specific failure modes, such as disturbances \cite{joshi2019deep, richards2021adaptive, hovakimyan20111}, incorrect models/parameters \cite{ljung1998system, kumar2021rma, brunton2016sparse}, or poor controller tuning \cite{loquercio2022autotune, marco2016automatic, berkenkamp2016safe}. However, these approaches work best under a pre-defined set of failure modalities, and/or leverage accurate models/prior knowledge about the robot from the designer.
Similarly, complex missions for autonomous mobile robots have been successfully managed through sophisticated state machines and mission planners \cite{furda2009towards,tranzatto2022team,agha2021nebula,agha2014firm,omidshafiei2015decentralized,bouman2020autonomous}. However, these planners often need to reason over a pre-defined set of states and/or observation models, identified through extensive efforts.  

\begin{figure}[t]
  \vspace{0.5em}
  \centering
    \includegraphics[width=\columnwidth, trim={0 0.6cm 0 0}, clip]{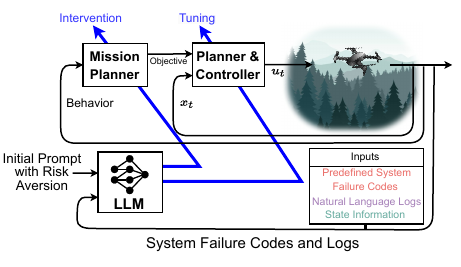}
    \caption{
        \footnotesize
        Schematic representation of our approach. We use a Large Language Model (LLM) to achieve adaptation from low-level to mission-scale decision-making, enabling resilience across the different dynamics and components of our autonomous system. Despite the ability of the LLM to apply changes to the low-level control of the multirotor, the LLM does not need to know specific details about the autonomy stack of the platform, instead using embedded prior knowledge to make decisions. In experiments, we demonstrate that an LLM queried from onboard the robot can reason about the current and desired state of the robot, deciding which corrective actions to apply to the control input to achieve the desired objective.
    }
    \label{fig:teaser}
    \vspace{-2em}
\end{figure}

Recently, foundational models, and especially Large Language Models (LLMs) pre-trained on internet-scale datasets \cite{openai2023gpt4, touvron2023llama1, touvron2023llama2}, have demonstrated impressive performance on a variety of reasoning problems, including natural language \cite{drori2022a, lewkowycz2022solving} and mathematics \cite{cobbe2021training}. This performance stems in part from the large size of their training data (e.g., internet-scale), which embeds a vast amount of prior knowledge into the weights of the model. Additionally, their billion-of-parameters model architectures enable reasoning over long sequences of symbols, causing them to be a natural choice for any problem that involves generating a sequence of symbols. The embedded prior knowledge and extended sequential reasoning capabilities have led to LLMs finding increased application in task planning and motion planning for robotics. In this context, the main focus of recent work has been planning for manipulation  \cite{saycan2022arxiv,liang2023code,huang2023instruct2act,wu2023tidybot,liu2023llmp,huang2022inner,huang2022language}, using human input as a task specification and outputting calls to manipulator APIs. However, their potential has not been explored for \textit{combined adaptive low-level control} and mission \textit{planning/reasoning} on agile autonomous aerial robots.

In this paper, we present REAL (Resilience and Adaptation using LLMs), a method for harnessing the capabilities of LLMs for mission-planning and low-level adaptive control of an agile mobile robot, a multirotor UAV. Our work leverages the LLM's embedded prior knowledge of the UAV's dynamics to create adaptation throughout the stack, including altering low-level parameters, producing commands to better track trajectories, and making mission-level decisions.
REAL uses a single human-crafted prompt (zero-shot prompting) to define minimal robot specification and task/controller API available to the LLM. Then, during real-time deployment, REAL receives as input a set of natural-language and numerical signals available onboard the multirotor which capture mission-relevant information at different timescales, including information about the dynamics of the robot and its high-level mission objectives. Then, based on these automatically generated robot prompts, REAL chooses the most suitable control/mission planning APIs that are executed by the robot. This feedback loop operates at about $0.1$-$1.0$ Hz, while the prompts are processed remotely using the \texttt{OpenAI GPT-4 API}.

We evaluate REAL in hardware experiments, exposing our multirotor to a variety of performance-lowering conditions. Some of these conditions require low-level adaptation (e.g., by adjusting the commanded thrust), while others which require mission-level adaptation (e.g., by improving controller tuning or conducting an emergency landing). 
Through these experiments, we also show that the behavior of an LLM as an adaptive controller can be modified by the use of natural language cues (e.g. using stronger language when the instructions given to the LLM are safety-related and important to follow).
Additionally, we show that although the LLM cannot be queried at a high rate (up to $1$ Hz), it can still process and make suggestions in response to high-frequency information by making use of algorithmically pre-processed information.

\textbf{Contributions}: %
\begin{itemize}
\item We present REAL, an approach to leverage prior knowledge in LLMs to enable online adaptation and decision-making across different time scales and components (\textit{low-level controller}, \textit{mission planner}) of the autonomy stack of an aerial robot. We leverage zero-shot prompting, and we show that our prompt requires minimal knowledge of the robot's model/dynamics and mission specifications. 
\item We present hardware experiments, demonstrating adaptation and decision-making capabilities using LLMs that improve the position control performance of the robot or regulate the safety of the mission. To the best of our knowledge, this is the first time that such capabilities have been demonstrated on an \textit{aerial} robot.  

\end{itemize}

\section{Related Works}

\noindent

\subsection{Adaptation at Mission-Scale and Low-Level Control}
\noindent
\textbf{Adaptive Control.} There are two broad categories of methods used for adaptive control: direct and indirect methods. Indirect methods aim at explicitly estimating models or parameters, which are leveraged in model-based controllers, such as MPC \cite{borrelli2017predictive}, to improve performance. Model/parameter identification include filtering techniques \cite{svacha2020imu, wuest2019online}, disturbance observers \cite{tagliabue2020touch, tagliabue2019robust, mckinnon2016unscented}, set-membership identification methods \cite{lopez2019adaptive, how2021performance} or learning-based methods \cite{saviolo2022active, kumar2021rma}. Direct methods, instead, develop policy updates that improve a certain performance metric. These updates are often done to drive the behavior of the system towards that of a reference model, with the updates themselves involving changing the shallow layers of the DNN policy \cite{joshi2019deep, joshi2020design, zhou2021bridging}. Other strategies include learning a policy update strategy offline using meta-learning \cite{richards2021adaptive, oconnel2022neural}, or using parametric adaptation laws such as $\mathcal{L}_1$ adaptive control \cite{hovakimyan20111}. While many adaptive control strategies are able to improve low-level performance in real-world systems, these strategies often fail when mission-level adaptation is required. Our work provides the first example of a system that exhibits both low-level and mission-level adaptation.

\noindent
\textbf{Uncertainty-Aware Mission Planning.} Mission-level adaptation is usually achieved with robot autonomy. State-of-the-art approaches to autonomy have involved the use of finite-state machines and uncertainty-aware planners \cite{omidshafiei2017decentralized, agha2014firm, bouman2020autonomous}, enabling autonomy on many systems, from a single autonomous car \cite{furda2009towards} to multiple heterogeneous robots \cite{tranzatto2022team, agha2021nebula}. %
While these methods achieve impressive performance in the coordination of multiple autonomous systems, they do not leverage the internet-scale prior knowledge in LLMs that may be helpful in making decisions under natural language-based observations that are available at the system level (e.g., log), nor they require to specify observations models/mission states.  

\subsection{Foundational Models in Robotics}

Foundational models have quickly found a variety of applications in robotics, with a focus on planning from natural language instructions. \cite{rt22023arxiv} develop a holistic foundational model that performs perception, planning, and control using internet-scale datasets to train a multi-modal foundational model that, given a goal described in natural language, can use video feed to plan and execute a sequence of commands to achieve that goal. \cite{saycan2022arxiv} decodes an LLM weighted by skill affordances \cite{zeng2019learning} from value functions to generate feasible plans for robots. \cite{liang2023code,huang2023instruct2act,wu2023tidybot} all translate a high-level instruction into a plan expressed in code, which is then executed by the robot. \cite{liu2023llmp} uses an LLM to translate a natural language planning problem into a domain-specific language, then runs a classical planner to solve the problem. \cite{huang2022language} uses an LLM to generate a plan in natural language, then uses a similarity measure to translate the plan from natural language into one executable by the robot. \cite{huang2022inner} uses closed-loop environmental feedback to improve the performance of using an LLM for planning and control in manipulation tasks.
While existing methods have focused on task-level planning (especially for manipulation), our work leverages LLMs for combined mission management and low-level control on an agile aerial robot, demonstrating a new domain of possible deployment of LLMs-based reasoning.

\section{Approach}

\subsection{Approach Overview}
The objective of our work is to design a decision-making and adaptation mechanism that uses LLMs to enable successful and resilient mission execution in autonomous systems despite the presence of uncertainties and potentially unplanned/unexpected failures that may happen across different levels of the autonomy stack. 
The considered autonomous system is a multirotor, whose objective consists in reaching and hovering at a desired position. During the mission, the robot is subject to uncertainties, such as model errors or wind, that may cause a critical mission failure. The robot needs to understand how to mitigate the effect of those uncertainties and autonomously decide whether to abort the mission if the effects of those uncertainties cannot be corrected, based on a natural-language specified risk tolerance. Our approach, summarized in \cref{fig:teaser}, leverages an LLM to trigger adaptive/resilient behaviors in the mission planning and control stack, taking as input available signals, pre-defined error codes, and natural-language based logs and error messages. In the following sections, we define in detail the interface between an existing autonomy stack and the LLM. 

\subsection{Autonomy Stack}
\subsubsection{Controller}
We consider a multirotor controlled by a cascaded position and attitude controller. The employed position controlled is based on a Linear-Quadratic Regulator (LQR) that uses a hover-linearized model (derived from \cite{kamel2017linear}) of an attitude-controlled robot of the form
$\mathbf{x}_{t+1} = \mathbf{A} \mathbf{x}_t + \mathbf{B} \mathbf{u}_t$, 
where $\mathbf{x} \in \mathbb{R}^{8}$ is the state and $\mathbf{u} \in \mathbb{R}^3$ is the control input. The state is
$\mathbf{x} = [
_W\mathbf{p}^\top, _W\mathbf{v}^\top, \prescript{}{I}\phi, \prescript{}{I}{\theta}
]^\top$, 
where $_W\mathbf{p}^\top \in \mathbb{R}^3$ and $_W\mathbf{v}^\top \in \mathbb{R}^3$ represent, respectively, the position and velocity expressed in a world frame $W$. The quantities $\prescript{}{I}{\phi}$ and $\prescript{}{I}{\theta}$ denote the attitude of the robot, expressed as roll and pitch Euler angles in a gravity-aligned, yaw-fixed frame $I$, whose x-axis is aligned with the world reference frame $W$.

We define the control input $\mathbf{u}$ to be:
$
\mathbf{u} = [ \prescript{}{I}{\phi}_\text{cmd}, \prescript{}{I}{\theta}_\text{cmd},  \delta f_\text{cmd}]^\top, 
$
where $\delta f_\text{cmd}$ denotes the linearized commanded thrust, and $\prescript{}{I}{\phi}_\text{cmd}$ and $\prescript{}{I}{\theta}_\text{cmd}$ are the commanded roll and pitch. These commands are executed by a cascaded attitude controller.

The control input is computed via:
\begin{equation}
\label{eq:lqr}
    \mathbf u_t = \mathbf \usafe_{t} + \mathbf K (\mathbf{x}_t - \vbf{x}^\text{ref}_{t}) + \delta \mathbf{u}_t,
\end{equation}
where $\mathbf \usafe_{t}$ represents the nominal command at hover and $\vbf{x}^\text{ref}_t$ a desired reference trajectory computed by the mission-level planner. $\mathbf K$ is a linear gain matrix, obtained by solving the Discrete Algebraic Riccati Equation (DARE) using the linearized model $\mathbf{A}$, $\mathbf{B}$ and given positive-definite tuning matrices $\mathbf R$ and $\mathbf Q$.
Key to this work, the additive control input $\delta \mathbf{u}_t$ represents an adaptive term that will be controlled by the LLM based on descriptions of the state of the system (error codes, logs), enabling adaptation at low-level control.

\subsubsection{Mission Planner}
The mission is managed by a finite state machine (FSM) that contains desired initial/terminal position setpoints, and timed transitions in between the desired states. Once a desired position is selected, the FSM generates reference trajectories (position, velocity) that are tracked via the position controller \cref{eq:lqr}. Every state inside the FSM is connected to an \texttt{emergency\_landing} action that leads to a safe state (on the ground below the robot), which can be triggered by the LLM upon seeing what it determines is sufficient cause to terminate the mission.

\subsection{Prompt Design And Interface with the Autonomy Stack}
In this section, we present the strategy to interface the control and mission/trajectory planning stack with the LLM. We use an approach inspired by \cite{liang2023code}, i.e., we leverage Python-based syntax to define the possible failure modes in the autonomy stack, as well as the description of a set of function callbacks (API) in our control framework available to execute corrective actions. However, our approach differs from \cite{liang2023code}, as we provide additional natural language instructions to express mission-level goals and trade-offs, i.e., the willingness to risk to continue the mission when complications arise, versus aborting the mission. Additionally, we limit the potentially dangerous execution of automatically generated Python code by providing the LLM with the instruction to call a set of pre-defined Python APIs. Last, in our experiments, the LLM is connected in a closed feedback loop with the rest of the autonomy stack, without human intervention beyond the initial prompt design.

\textbf{Code Color Convention:} Note that throughout this work we use the following convention: \query{green} denotes the initial prompt to the LLM; this prompt is hand-crafted by a human and is loaded at the start of the mission; \prompt{grey} denotes the query automatically generated by the autonomous system, and \resp{blue} denotes the reply from the LLM, closing the feedback loop.

Our prompt begins with the following sentence:
\lmpchat{Initial Prompt (Part 1)}{
\query{
\# Inside the codebase of my multirotor I found the following python code:
}%
}
This sentence introduces the LLM to the Python-based syntax that will be used next to list possible mission failures/issues, requirements, and actions available, and additionally introduces the LLM to the type of platform it needs to control. 
Next, we introduce a list of possible, easy-to-monitor, state-based errors and failures:
\lmpchat{Initial Prompt (Part 2)}{
\query{
\# list of possible issues/failures in mission planner/controller:\\
NO\_ISSUE = 0\\
FLYING\_TOO\_HIGH = 3\\
FLYING\_TOO\_LOW= 4\\
FLYING\_TOO\_LARGE\_POSITIVE\_POSITION\_ERROR\_X = 7\\
FLYING\_TOO\_LARGE\_NEGATIVE\_POSITION\_ERROR\_X = 8\\
FLYING\_TOO\_LARGE\_POSITIVE\_POSITION\_ERROR\_Y = 5\\
FLYING\_TOO\_LARGE\_NEGATIVE\_POSITION\_ERROR\_Y = 6
}%
}%

These failures can be easily detected, and their corresponding number is fed as input to the LLM. Additionally, we found that the LLM is more easily able to interpret failures expressed in natural language than failures expressed in numerical signals (i.e., current trajectory tracking errors).
The corresponding error codes are generated by comparing the current trajectory tracking error $\mathbf{p}_t - \mathbf{p}^\text{des}_t$, and by triggering an issue on the corresponding axis if the error exceeds a predefined threshold.

Next, we define a new fictitious Python variable and function call that computes the possible failures:
\lmpchat{Initial Prompt (Part 3)}{
\query{
\# check current failure using check\_failure. outputs a list of possible failures, for example [2, 3], \\ 
\# and a string with additional information. The string may be empty. \\
\# Example current\_failure: ([2, 3], 'position error = [0.1, -0.1, 1.5]') \\
current\_failures = check\_failures()
}%
}

As in \cite{liang2023code}, we make use of Python comments to provide contextual information on the output of the function call and describe additional inputs that we will be feeding into the LLM, using the second term in tuple of \texttt{current\_failure}. This extra input can be used to provide descriptive error messages or other information that is not known/does not need to be specified a priori, providing additional flexibility in the type of information that we can feed to the LLM.  

Next, we provide the LLM with information about the system-level actions (APIs) that the LLM can select: 

    \lmpchat{Initial Prompt (Part 4)}{
    \query{
    \# possible failure mitigation strategies\\
    from controller import (\\
    \hspace*{0.2cm}\# modify control input\\
    \hspace*{0.2cm}increase\_thrust, decrease\_thrust, accel\_positive\_x, accel\_negative\_x, accel\_positive\_y, accel\_negative\_y,\\
    \hspace*{0.2cm}\# Mission-level decisions\\
    \hspace*{0.2cm}emergency\_landing, do\_nothing,\\
    \hspace*{0.2cm}\# Controller tuning -- we use a LQR\\
    \hspace*{0.2cm}tune\_controller\_by\_decreasing\_the\_cost\_of\_actuation\_usage,\\
    \hspace*{0.2cm}tune\_controller\_by\_increasing\_the\_cost\_of\_actuation\_usage,\\
    \hspace*{0.2cm}tune\_controller\_by\_increasing\_penalty\_on\_position\_errors,\\ 
    \hspace*{0.2cm}tune\_controller\_by\_decreasing\_penalty\_on\_position\_errors,\\
    )\
    }%
    }%

These actions correspond to changes in the control inputs or to events in the mission planner. 
More specifically, \texttt{increase\_thrust} and \texttt{decrease\_thrust} increase/decrease an adaptive term in the control input, while \texttt{accel\_positive\_...} and \texttt{accel\_negative\_...} produce accelerations along an axis by increasing/decreasing the extra roll/pitch setpoints by a pre-specified amount. Additionally, \texttt{tune\_controller\_...} updates the corresponding part of the weight matrices $\mathbf{R}$ and $\mathbf{Q}$ of the position controller \cref{eq:lqr}; the corresponding DARE is solved onboard and the resulting gain matrix updates $\mathbf{K}$.

Towards the end of the prompt, we switch back to natural language to provide mission specifications: 
\lmpchat{Initial Prompt (Part 5)}{
\query{
From now on, I provide you with the value of the variable “current\_failure”,\\ 
and your output needs to be your best guess of the function names in the python list\\
"list\_of\_function\_names\_to\_be\_executed\_right\_now".\\
For instance, your output: ["emergency\_landing"],"low\_battery\_voltage", \\
"because the drone can hardly move it is safer to land"\\
Try to think like a drone control engineer.
}%
}

This prompt specifies the output that we expect from the LLM (a list of names of functions the controller can execute). It additionally includes two elements that can help the LLM reason about its choice of actions, and a brief and long explanation of the issue. Following best prompting practices, we also encourage the LLM to role-play, i.e., thinking like a "drone control engineer". 

In addition, we discourage the LLM from outputting planned future actions and encourage brevity in its explanations, by adding the following lines in the initial prompt:
\lmpchat{Initial Prompt (Part 6)}{
\query{
DO NOT output function names to be called in the future, but account for past problems to come up\\
with your guess of the functions in "list\_of\_function\_names\_to\_be\_executed\_right\_now".
}%
}

Last, we further make the LLM aware of the possibility of taking emergency landings: 
\lmpchat{Initial Prompt (Part 7)}{
\query{
If problems persist, do not hesitate to emergency land.\\
if your actions do not take the desired effect, you must perform an emergency landing.
}
}
We note that omitting these sentences was making the LLM less prone to trigger an emergency landing, while exaggerating the need to emergency land (e.g., using "MUST" instead of "must") made the LLM more prone to immediately trigger an emergency landing, potentially providing a natural-language avenue to specify willingness to take risks in an autonomous system.

\section{Evaluation}
\subsection{Implementation Details}

We perform real-world experiments by deploying REAL on a multirotor. The multirotor is equipped with an Intel\textsubscript{\textregistered} NUC\textsuperscript{TM} 10. Our system operates in real-time on the NUC\textsuperscript{TM} 10, driven by the Intel\textsubscript{\textregistered} Core\textsuperscript{TM} i7-10710U Processor. All planning, control, and state estimation, which merges IMU data with a motion capture system, are executed onboard, except the LLM, which is queried from onboard, receiving replies generated via the \texttt{OpenAI GPT-4 API}. The UAV connects to the internet over Wi-Fi and queries the LLM at as high of a rate as possible. This results in the LLM running at from $0.1$ to $1.0$ Hz, depending on network latency and API usage.

\subsection{Low-Level Adaptation and Controller Auto-Tuning}

In this experiment, we evaluate the ability of the LLM to perform low-level adaptive control and decision-making, by adjusting the control input of our UAV based on the issues reported to the LLM. To introduce tracking error, we purposefully use the wrong value for the mass parameter while synthesizing the controller; the value of the mass parameter used is about $15\%$ of the robot's true mass, resulting in a large altitude error.
We then deploy the UAV in a mission that consists of taking off, following a figure-eight trajectory, and then landing. 
We repeat the experiment two times, with the difference that in the first experiment, to study the choice that the LLM would make absent this parameter, we removed the \texttt{tune\_controller\_by\_..} in \texttt{Initial Prompt (Part 4)} that API call.

\begin{figure*}[h!]
    \centering
    
    \begin{subfigure}{0.48\textwidth}
    \centering
       \includegraphics[trim={5in 1in 0 0},clip,width=\textwidth]{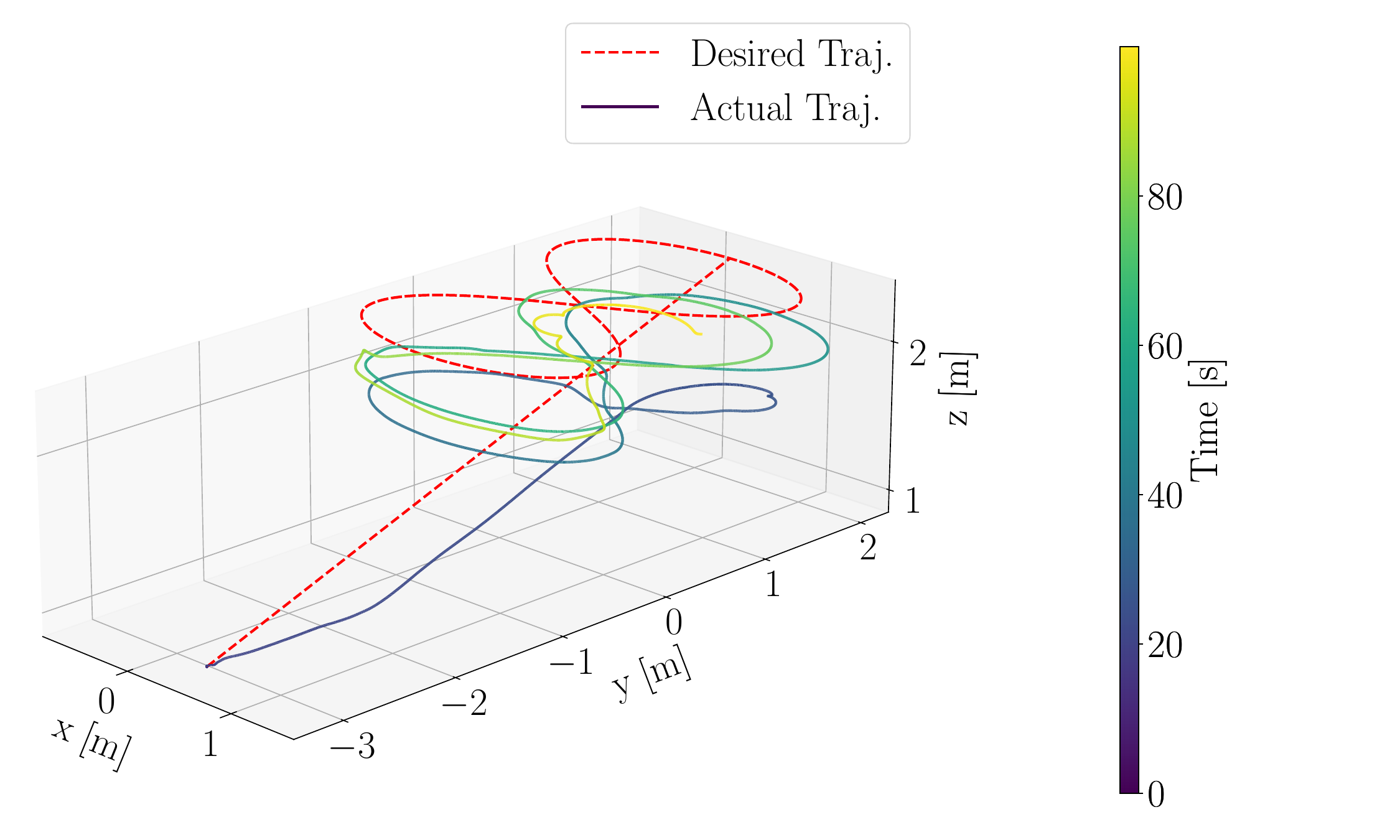}
    \caption{Trajectory Tracking \textbf{without} Control Gain $\mathbf{K}$ Tuning}
    \label{fig:hw-success-wrong-mass-with-llm-no-gain-tuning}
    \end{subfigure}%
    \vspace{0.1cm}
    \begin{subfigure}{0.48\textwidth}
    \centering
    \includegraphics[trim={5in 1in 0 0},clip,width=\textwidth]{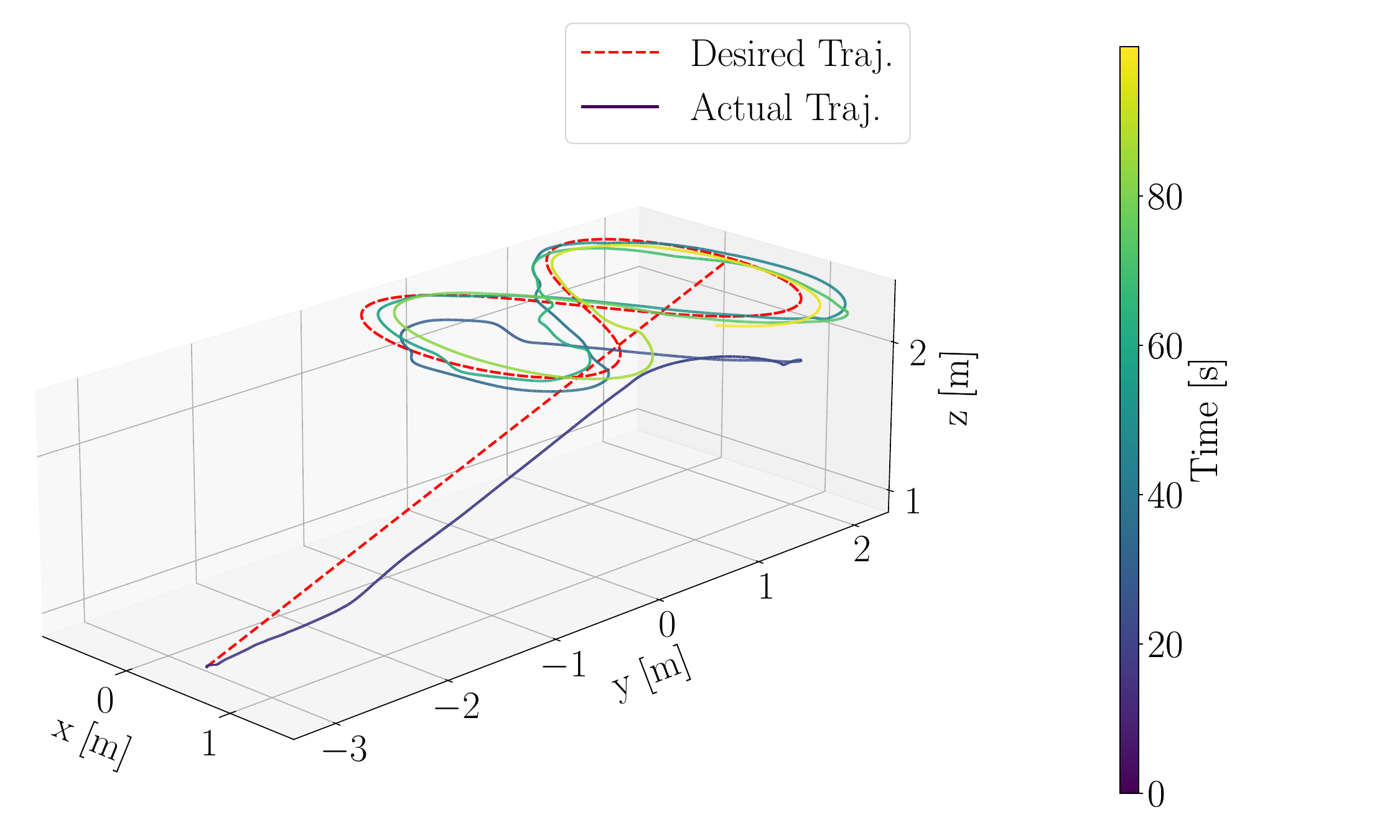}
    \caption{Trajectory Tracking \textbf{with} Control Gain $\mathbf{K}$ Tuning.}
    \label{fig:hw-success-wrong-mass-with-llm-with-gain-tuning}
    \end{subfigure}
        
        \caption{   
    REAL's successful altitude ($z$-axis) adaptation during a figure eight trajectory (\SI{0.25}{\m/\s}) with an 
    incorrect mass parameter used in the controller (\SI{15}{\%} error from nominal value). 
    The top figure shows the experiment performed when the API calls \texttt{tune\_controller\_by\_...} removed from the initial prompt, and the bottom figure shows the same experiment with the \texttt{tune\_controller\_by\_...} APIs added back. These results illustrate that REAL can successfully interpret the system-provided error messages, and call useful APIs. Specifically, REAL achieves adaptation in the top plot by calling the \texttt{increase\_thrust} API, successfully converging to the desired altitude. In the secondary scenario (bottom figure), REAL calls a combination of \texttt{tune\_controller\_by\_...} and \texttt{increase\_thrust} APIs, triggering changes in the controller matrix $\mathbf{K}$. This results in faster/better convergence to the desired trajectory. 
    }
    \label{fig:hw-success-wrong-mass-with-llm}
\end{figure*}

\cref{fig:hw-success-wrong-mass-with-llm-no-gain-tuning} shows the result without tuning the API call \texttt{tune\_controller\_by\_..} in \texttt{Initial Prompt (Part 4)}, highlighting that throughout the mission, REAL succeeds at improving the altitude improve the trajectory tracking through the duration of the experiment (about \SI{100}{\s}) by repeatedly calling the \texttt{increase\_thrust} in \texttt{Initial Prompt (Part 4)}, obtaining an altitude tracking error within \SI{30}{\cm}.
\cref{fig:hw-success-wrong-mass-with-llm-with-gain-tuning} shows the results on the same trajectory, but with the \texttt{tune\_controller\_by\_..} in \texttt{Initial Prompt (Part 4)} added back to the prompt. In this experiment, the LLM calls both commands, as shown by the conversation between the LLM and the robot during the experiment (please see Conversation 1 in the Appendix).

The altitude tracking error resulting from this experiment is within only $10$ cm. This shows that the LLM is capable of selecting multiple mission-relevant low-level control APIs, without providing detailed prior instructions on its choice, and whose usage is beneficial at improving the outcome of the mission.

\subsection{Low-Level Adaptation to Unmodeled Dynamics}
\begin{figure}
    \centering
    \begin{subfigure}{.5\textwidth}
        \centering
        \includegraphics[width=\textwidth, trim={0.5cm 0.18in 0 0}, clip]{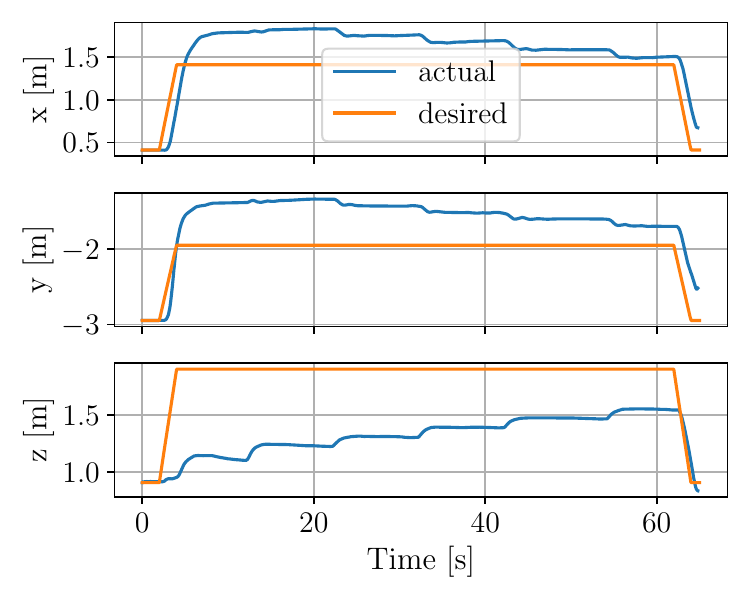}
        \caption{REAL reducing position errors across the $x$, $y$, and $z$-axes under unmodeled dynamics: In this experiment an additional weight of \SI{210}{\g} was added to one of the UAV's arms, creating a large unmodeled external torque that affects the attitude and position control of the UAV. REAL identifies the correct APIs to reduce this error, though convergence is slowed down due to the limited rate at which the API can be called.}
        \label{fig:mass-on-arm}
    \end{subfigure}%
    \hfill %
    \begin{subfigure}{.45\textwidth}
        \centering
            \begin{tikzpicture}[every text node part/.style={align=center}]
        \node {\includegraphics[trim={0.2in 0.2in 0 0},clip,width=0.8\textwidth]{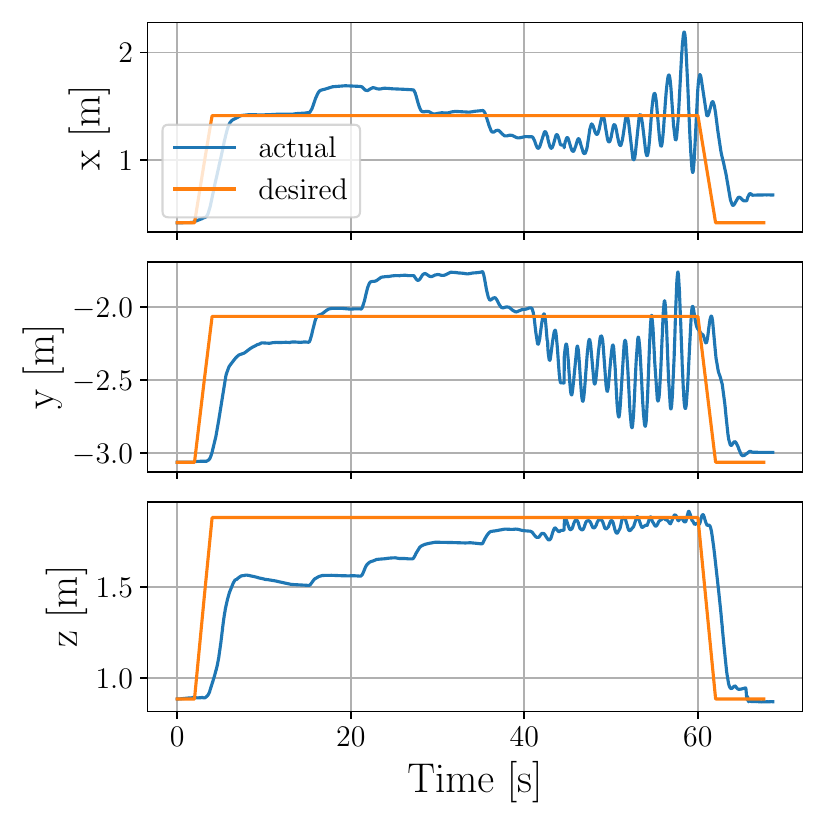}};
    \node at (1.8, 0.2) [minimum width=0.001in, minimum height=1.9in, fill=red, opacity=0.5] {};
    \node at (2.2, -2.5) [rectangle, red] {\scriptsize \query{emergency\_landing} called};
    \end{tikzpicture}
        \caption{REAL's position adaptation combined with its ability to trigger emergency landing automatically. The UAV was configured with a \SI{15}{\%} mass-mismatch, that is successfully compensated by REAL in the initial phase of the experiment. Subsequently, we artificially introduced oscillations by pulling a cable attached to the UAV. Upon receiving a natural-language log that informs REAL of these oscillations, the LLM invokes the \texttt{emergency\_landing} API. This experiment highlights REAL's ability to handle both lower-level control adaptations and higher-level mission-critical decisions such as aborting the mission for safety reasons.}
        \label{fig:oscillations}
    \end{subfigure}
    \caption{Real's ability to perform low-level adaptation and trigger mission-relevant decision. }
\end{figure}

This experiment presents a more challenging scenario that highlights REAL's ability to correct for unmodeled dynamics. 
To test this, we place a large extra mass on the end of one of the multirotor's arms, creating an unmodeled torque disturbance. For brevity, the prompt is not shown, but adaptation is triggered by repeatedly selecting the expected API calls: \texttt{increase\_thrust}, \texttt{accel\_negative\_y} and \texttt{accel\_negative\_x}.  
Fig.~\ref{fig:mass-on-arm} shows the hardware experiment results of the LLM successfully reasoning how to eliminate error along each of its axes. 
Note that in this earlier experiment, roll and pitch torque function names were used in the prompt to control acceleration along the $y$ and $x$ axes respectively. We later found that the LLM was more consistent when using the commands for requesting acceleration in $x$ and $y$ directions directly.

\subsection{Mission-scale Decision Making via Unsafe State Detection and Automatic Mission Abortion}
The purpose of this experiment is two-fold: 
(1) test the LLM's ability to make critical mission-level decisions in the event that unforeseen circumstances cause the UAV to lose control
and 
(2) test the LLM's ability to process additional information that was not in the original prompt.

Since the LLM is only able to make adaptive corrections at a slow rate, we would expect it to call for an emergency landing in the event of loss-of-control.
To simulate a loss-of-control event, we apply large external disturbances to the robot by pulling it via a rope, causing large oscillations along the $y$ axis, as seen in \cref{fig:oscillations}.
Onboard the drone, we implement a module that stores a buffer of positions and uses a Fast Fourier Transform (FFT) to evaluate the maximum amplitude of any frequency content. If the maximum frequency content is above a threshold, this information is outputted to the log received by the LLM, as shown in the conversation history in Conversation 2 (please see Conversation 2 in Appendix). Although the LLM was not told it would receive any information about oscillations nor what to do if it did receive this information, it was able to make the critical decision to select the \texttt{emergency\_land} API when it recognized that the UAV was in a dangerous condition via the provided log message "\texttt{DANGEROUS oscillations...}". \cref{fig:oscillations} additionally highlights the LLM's ability to correct the altitude errors of the UAV, caused by an additional $15\%$ mass mismatch, as visible before the interaction begins.

\section{Conclusions}
We have presented REAL, a method to utilize LLMs for zeros-shot cross-stack adaptation and autonomy on mobile robots. The method works by leveraging LLMs' natural language understanding, its ability to reason over long sequences, and its embedded prior knowledge of the robot's model/dynamics. Our evaluation in hardware experiments has demonstrated promising first-steps towards low-level adaptation and increased mission-level resilience. In the future we would like to further extend this framework to more complex, multi-robot autonomous systems, where more complex reasoning is required to diagnose and recover from failures.

\clearpage

\bibliography{ref}  %

\section{Appendix}
\subsection{Conversation 1}
\lmpchat{Conversation 1}{
\query{
}\\
\hlcoderesp{t = 3.92s: \prompt{Prompt ([0], '')}}\\
\hlcoderesp{t = 3.92s: \resp{Response list\_of\_function\_names\_to\_be\_executed\_right\_now: ["do\_nothing"], "no\_issue", "Since there }}\\
\hlcoderesp{\resp{\Repeat{12}{\ }are currently no issues detected with the drone, no corrective actions are needed at this time." }}\\
\hlcoderesp{t = 8.47s: \prompt{Prompt ([4, 6], 'z error is -0.14, y error is -0.44, ')}}\\
\hlcoderesp{t = 8.47s: \resp{Response list\_of\_function\_names\_to\_be\_executed\_right\_now: ["increase\_thrust", "accel\_positive\_y"] }}\\
\hlcoderesp{\resp{\Repeat{12}{\ }"flying\_too\_low\_and\_negative\_y\_position", "The drone is currently flying too low and also has a }}\\
\hlcoderesp{\resp{\Repeat{12}{\ }negative error in the Y-axis. Increasing thrust will correct the altitude issue, and accelerating in }}\\
\hlcoderesp{\resp{\Repeat{12}{\ }the positive Y direction will correct the positional error."}}\\
\hlcoderesp{t = 14.03s: \prompt{Prompt ([4, 6], 'z error is -0.63, y error is -1.18, ')}}\\
\hlcoderesp{t = 14.03s: \resp{Response list\_of\_function\_names\_to\_be\_executed\_right\_now: ["increase\_thrust", "accel\_positive\_y",  }}\\
\hlcoderesp{\resp{\Repeat{13}{\ }"tune\_controller\_by\_increasing\_penalty\_on\_position\_errors"], "flying\_too\_low\_and\_large\_negative}}\\
\hlcoderesp{\resp{\Repeat{13}{\ }\_y\_position", "The drone is flying significantly too low and has a large negative error on the Y-axis.}}\\
\hlcoderesp{\resp{\Repeat{13}{\ }It's necessary to increase thrust and accelerate in the positive Y direction to correct these issues. }}\\
\hlcoderesp{\resp{\Repeat{13}{\ }Additionally, the large errors suggest it may be beneficial to tune the controller to penalize}}\\
\hlcoderesp{\resp{\Repeat{13}{\ }position errors more severely."}}\\
\hlcoderesp{t = 19.0s: \prompt{Prompt ([4, 5, 7], 'z error is -0.30, y error is 0.15, x }}\\
\hlcoderesp{\prompt{\Repeat{12}{\ }error is 0.12, ')}}\\
\hlcoderesp{t = 19.0s: \resp{Response list\_of\_function\_names\_to\_be\_executed\_right\_now: ["increase\_thrust", "accel\_negative\_y",}}\\
\hlcoderesp{\resp{\Repeat{12}{\ }"accel\_negative\_x"], "flying\_too\_low\_and\_positive\_position\_errors", "The drone is below the desired }}\\
\hlcoderesp{\resp{\Repeat{12}{\ }altitude and has minor positive errors in both X and Y axis. To correct these, it should increase }}\\
\hlcoderesp{\resp{\Repeat{12}{\ }thrust to gain altitude, and decelerate in Y and X directions."}}\\
}

\subsection{Conversation 2}
\lmpchat{Conversation 2}{
\query{
}\\
\hlcoderesp{t = 6.07s: \prompt{Prompt ([0], '')}}\\
\hlcoderesp{t = 6.07s: \resp{Response list\_of\_function\_names\_to\_be\_executed\_right\_now: ["do\_nothing"]
reason: The reported }}\\
\hlcoderesp{\resp{\Repeat{11}{\ }
information indicates that there are currently no discernible issues with the drone. Therefore, no }}\\
\hlcoderesp{\resp{\Repeat{11}{\ }
actions are necessary at this time.}}\\
\hlcoderesp{t = 13.95s: \prompt{Prompt ([4, 6, 8], 'z error is -0.69, y error is -0.62, 
}}\\
\hlcoderesp{\prompt{\Repeat{12}{\ }
x error is -0.40, ')}}\\
\hlcoderesp{t = 13.95s: \resp{Response list\_of\_function\_names\_to\_be\_executed\_right\_now: ["increase\_thrust", "tune\_controller\_by}}\\
\hlcoderesp{\resp{\Repeat{12}{\ }
\_decreasing\_the\_cost\_of\_actuation\_usage", "accel\_positive\_y", "accel\_positive\_x"] reason: The drone 
}}\\
\hlcoderesp{\resp{\Repeat{12}{\ }
is flying too low and has negative position errors in both X and Y directions, hence, increasing 
}}\\
\hlcoderesp{\resp{\Repeat{12}{\ }
thrust is the first step, along with acceleration in positive X and Y directions to correct the 
}}\\
\hlcoderesp{\resp{\Repeat{12}{\ }
position errors. Besides, the cost of actuation usage should also be decreased to ensure a higher 
}}\\ 
\hlcoderesp{\resp{\Repeat{12}{\ }
rate of control input changes for faster response. It's preferable to decrease the cost of 
}}\\
\hlcoderesp{\resp{\Repeat{12}{\ }
actuation usage to make the drone respond more swiftly to the control inputs, as there are multiple  
}}\\
\hlcoderesp{\resp{\Repeat{12}{\ }
issues happening simultaneously.}}\\
\hlcoderesp{t = 57.48s: \prompt{Prompt ([6, 8], 'y error is -0.65, x error is -0.28, VERY 
}}\\
\hlcoderesp{\prompt{\Repeat{12}{\ }
DANGEROUS oscillations on y-axis. Frequency is 0.67 [Hz], 
}}\\
\hlcoderesp{\prompt{\Repeat{12}{\ }
amplitude is 0.19 [m].')}}\\
\hlcoderesp{t = 57.48s: \resp{Response list\_of\_function\_names\_to\_be\_executed\_right\_now: ["emergency\_landing"] reason: The drone 
}}\\
\hlcoderesp{\resp{\Repeat{12}{\ }
has large errors  in both X and Y directions, and additionally, is exhibiting dangerous oscillations 
}}\\
\hlcoderesp{\resp{\Repeat{12}{\ }
on the Y-axis. This indicates unstable flight dynamics which could be detrimental to the safety of the
}}\\
\hlcoderesp{\resp{\Repeat{12}{\ }
operation. An immediate emergency landing should be executed to prevent potential damage or hazards.
}}\\
}

\end{document}